\documentclass{article}
\usepackage{spconf,amsmath,epsfig}
\usepackage{subfigure}
\usepackage{bbding}

\title{HYBRID FEATURE EMBEDDING FOR AUTOMATIC BUILDING OUTLINE EXTRACTION}
\name{Weihang Ran\textsuperscript{1}, Wei Yuan\textsuperscript{1*}
\thanks{*Corresponding author}
, Xiaodan Shi\textsuperscript{1}, Zipei Fan\textsuperscript{1}, Ryosuke Shibasaki\textsuperscript{1}}
\address{\textsuperscript{1}Center for Spatial Information Science, the University of Tokyo\\
    \{ranweihang, miloyw, shixiaodan, fanzipei, shiba\}@csis.u-tokyo.ac.jp
    }
\begin{document}
%
\maketitle
\begin{abstract}
Building outline extracted from high-resolution aerial images can be used in various application fields such as change detection and disaster assessment. However, traditional CNN model cannot recognize contours very precisely from original images. In this paper, we proposed a CNN and Transformer based model together with active contour model to deal with this problem. We also designed a triple-branch decoder structure to handle different features generated by encoder. Experiment results show that our model outperforms other baseline model on two datasets, achieving $91.1\%$ mIoU on Vaihingen and $83.8 \%$ on Bing huts. 
\end{abstract}
\section{Introduction}
\label{sec:1}
In recent years, semantic segmentation research base on high-resolution aerial images has achieved great progress, but it's more convenient to use contour information of target objects instead of pixel-wised segmentation result in practical data storage and application scenarios. Therefore, object contour extraction is a very important research task in downstream applications. However, research aiming at extracting contours from aerial imagery is very limited, which motivated us to carry out this research on using deep learning technology to obtain contour information from high-resolution aerial imagery.

In the view of the powerful ability of convolutional neural network to process image data, researchers first applied various well-performing CNN models to the processing of aerial images to extract buildings.
These models can assign a class label to a single pixel in the image, thus achieving the separation of the object from the background, or the distinction between different categories. But they cannot distinguish different objects in the same category, and the outlines generated by CNN are always unsatisfactory.

Different from the traditional perspective of semantic segmentation, active contour model uses pixel value information of image to generate contour in the way of curve evolution. So it is considered to have the potential to solve this problem. The first active contour model, which is known as snakes \cite{ref5}, defined an energy functional and found the boundary of different objects in the image by minimizing this target functional. Some research attempted to construct a loss function in the similar form of it to enhance the image segmentation performance \cite{ref6}. DSAC \cite{ref7} applied CNN to learning the energy landscape and combined the results with snakes, achieved end-to-end building contour extraction as a result. Instead of using snakes, DARNet \cite{ref8} employed a reformulated active contour model named active rays \cite{ref9} together with dilated residual network \cite{ref10} to learn energy landscapes. Although these methods have made some progress in the contour extraction task, the potential of the active contour model has not been fully explored due to the limitation of the pure convolution-structured encoder. With the emergence of Transformer \cite{ref11} and Vision Transformer \cite{ref12}, self-attention mechanism has shown its ability to extract long distance dependencies. So it has attracted many researchers to combine it with CNN to explore its potential in the field of computer vision \cite{ref13, ref14}. Similarly, the combination of self-attention and active contour models has a very broad prospect.

So in this paper, we proposed a CNN and Transformer based model for automated building contour extraction. The contributions can be summarized as follows: 1) we proposed a hybrid model for building extraction which outperforms other baseline models on two open-source datasets; 2) we designed a triple-branch decoder to deal with different features output by different encoder part. The details of our model is introduced in Section \ref{sec:2}. Our experiment results are shown in Section \ref{sec:3}. Section \ref{sec:4} conclude this paper.


\section{Methodology}
\label{sec:2}
\begin{figure*}
    \centering
    \includegraphics[width=6.0in]{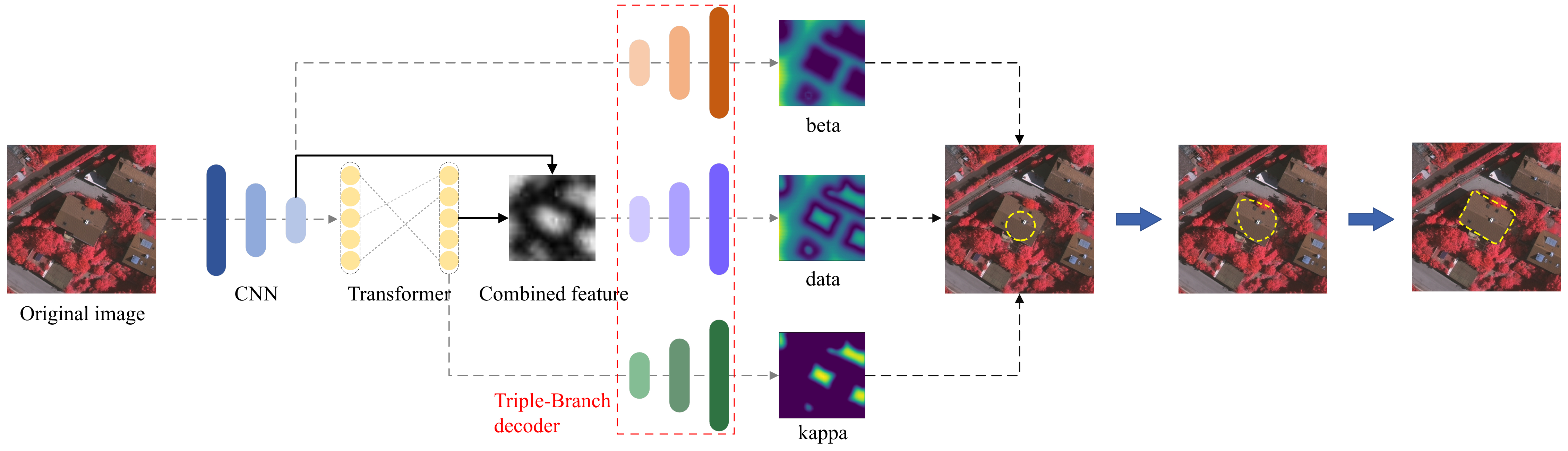}
    \caption{Architecture of our model. The original image is first processed by a CNN then a Transformer module. The extracted features will be upsampled by three different branches and generate corresponding data, beta and kappa.}
    \label{fig:model}
\end{figure*}

As is shown in Figure \ref{fig:model}, given an image with shape of $[3, H, W]$ as input, our model will first extract locality information by doing convolution operation. Then the high dimensional middle features will be transferred into linear tensors and processed by multiple self-attention blocks. In order to take full advantage of the various features, three branches are adopted to upsample them respectively to obtain different energy landscapes in the active contour model. Finally the active contour model is applied to evolve the initial contour into precise one by minimizing energy functional.

\subsection{Hybrid encoder}
\label{ssec:encoder}
The encoder-decoder structure has been proven effective in many classic computer vision research. CNN is considered a powerful encoder in the past. Although with the birth of Transformer, the limitations of its ability to construct long-distance relationships are magnified, we still believe that CNN has irreplaceable advantages in constructing local features. Therefore, the encoder part of our model used a hybrid architecture of dilated residual network and Vision Transformer to give full play to the different advantages of these two operations.

\subsection{Triple branch decoder}
\label{ssec:decoder}
\begin{figure}[!htp]
    \centering
    \begin{minipage}[b]{0.27\linewidth}
        \centering
        \subfigure{\includegraphics[width=0.75in]{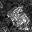}}
        \subfigure{\includegraphics[width=0.75in]{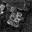}}
        \subfigure{\includegraphics[width=0.75in]{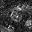}}
        \centerline{(a)}
    \end{minipage}
    \begin{minipage}[b]{0.27\linewidth}
        \centering
        \subfigure{\includegraphics[width=0.75in]{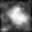}}
        \subfigure{\includegraphics[width=0.75in]{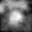}}
        \subfigure{\includegraphics[width=0.75in]{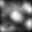}}
        \centerline{(b)}
    \end{minipage}
    \begin{minipage}[b]{0.27\linewidth}
        \centering
        \subfigure{\includegraphics[width=0.75in]{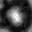}}
        \subfigure{\includegraphics[width=0.75in]{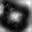}}
        \subfigure{\includegraphics[width=0.75in]{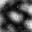}}
        \centerline{(c)}
    \end{minipage}
    \caption{Visualization of intermediate features. (a), (b), (c) are the features output by CNN, Transformer and their combination respectively.}
    \label{fig:features}
\end{figure}

From the intermediate features generated by CNN and Transformer in Figure \ref{fig:features}, we can see that CNN focuses more on the boundaries of objects while Transformer is more efficient in extracting the main body of the target object. This means that they are suitable for calculating different data terms in active contour model. And if we combine these features together, more refined features can be generated. Encouraged by this discovery, we designed a triple-branch decoder structure containing three upsampling process to deal with different features. The intermediate features have the same shape of $[512, H/8, W/8]$. After upsampling they will be converted into three one-dimensional energy landscapes with the same size as original image. 

\subsection{Active contour}
\label{ssec:acm}
Given a binary image, active contour model can find the demarcation between object and background automatically and recognize it as outline. It is achieved by defining an energy functional and minimizing it using the method of curve evolution. In order to simplify the computation, we discretize the contour as a sequence of $\emph{L}$ vertexes $\{ c_{i} \} ^{\emph{L}}_{i=1}$ and represent them in a polar coordinates way just like in DARNet:
\begin{equation}
c_{i} = \left[ 
\begin{array}{c}
x_{c} + \rho_{i}cos(i\Delta\theta) \\
y_{c} + \rho_{i}sin(i\Delta\theta) \\
\end{array}
\right]
\end{equation}

The energy functional is defined as follows:
\begin{equation}
\label{eq:energy}
\emph{E}(c) = \sum^{\emph{L}}_{i=1} \left[ \emph{E}_{data}(c_{i}) + \emph{E}_{curve}(c_{i}) + \emph{E}_{balloon}(c_{i}) \right]
\end{equation}
where $\emph{E}_{data}$, $\emph{E}_{curve}$, $\emph{E}_{balloon}$ refer to data term, curvature term and balloon term. By our deep learning model, we can learn $\emph{D}$, $\beta$ and $\kappa$ from original image in feature map forms. Then $\emph{E}_{data}$, $\emph{E}_{curve}$, $\emph{E}_{balloon}$ will be calculated in the following way:
\begin{equation}
\label{eq:data}
\emph{E}_{data}(c_{i}) = \emph{D}(c_{i})
\end{equation}

\begin{equation}
\label{eq:curve}
\emph{E}_{curve}(c_{i}) = \beta(c_{i}) |c_{i+1} - 2c_{i} + c_{i-1}|^{2}
\end{equation}

\begin{equation}
\label{eq:balloon}
\emph{E}_{balloon}(c_{i}) = \kappa(c_{i}) ( 1 - \rho_{i} / \rho_{max} )
\end{equation}

For data term, $\emph{D}$ is a non-negative feature map generated by our model, and $\emph{D}(c_{i})$ refers to the pixel value of point $c_{i}$ in this feature map. $\emph{D}$ is the basis of active contour model used to divide objects and background, so it should give low pixel values near the boundary theoretically.

For curvature term, $\beta$ is a non-negative feature map, while $c_{i-1}$ and $c_{i+1}$ are the last point and the next point of $c_{i}$. The squared term is a discrete approximation of the second order derivative of the contour. This is used to control the elasticity of the contour, deciding where to bend and where to straighten.

$\kappa$ in the balloon term is also a non-negative feature map output by our model. $\rho_{max}$ is the maximum radius a contour can reach without going beyond the boundaries of original images. This term is applied to control contour points within a reasonable range of reference points and complement the data term and curvature term.

In order to make the initial contour evolve to the optimal one, we calculate the partial derivative of each term and set them to zero. Then an iterative method is applied to solve the partial derivative system.



\section{Experiments}
\label{sec:3}

\subsection{Datasets}
\label{ssec:dataset}
In our experiment, we used two open-source building detection datasets.
The Vaihingen dataset contains 168 images with $512 \times 512$ size cropped from the ISPRS ``2D semantic labeling contest". Each image has a label file that contains only the mask of the target building and a label file that contains all the buildings. A csv file that contains the vertex of the target building contour is also included. We extract 140 images for training/validating and 28 images for testing.
The Bing nuts dataset is collected from Tanzania and consists of 606 images with $64 \times 64$ pixel size and 30 cm spatial resolution. The ground truth is got from OpenStreetMap. In our experiment we split the whole dataset into 500/106 images for training/testing respectively.

\subsection{Implementation details}
\label{ssec:setup}
In consideration of the datasets used in our experiment is insufficient to train a large model from the beginning, we used some pretrained parameters to accelerate training process. The backbone used in our model is dilated residual network followed by ViT-Base with 768 dimensional hidden states and 6 self-attention blocks. All of the checkpoints are pretrained on Imagenet21k datasets.

We first pretrain our model for 250 epochs applying the Stochastic Gradient Descent (SGD) optimizer with 0.9 momentum and a weight decay of $4 \times 10^{-4}$. The initial learning rate is $1 \times 10^{-3}$, reducing by half every 30 epochs. After pretraining process we train the model for 30 epochs using SGD optimizer with momentum of 0.3 and a weight decay of $1 \times 10^{-5}$. The initial learning rate will be halved every 50 epochs from $4 \times 10^{-5}$. Batch size is 8. The training process is conducted on a single NVIDIA GeForce RTX 3090 GPU.

\subsection{Result}
\label{ssec:result}
We compared the results of our model with DSAC and DARNet. All the experiment results are summarized in Table \ref{tab:baseline}. We used mean intersection over union (mIoU) to evaluate the performance of each model on Vaihingen dataset. However, the ground truth shift noise in the Bing huts dataset makes the IoU assessment untrustworthy, we also used the root mean square error (RMSE in $\mathit{m^{2}}$) to estimate the area that predicted correctly. From the table we can see that our TriTransCon outperform the other baselines in terms of mIoU on both datasets, especially on Bing huts. And the RMSE is also improved greatly by our model on Bing huts datasets.

\begin{table}[ht]
    \centering
    \begin{tabular}{|c|c|c|c|} 
        \cline{2-4}
        \multicolumn{1}{c|}{}    & \multicolumn{2}{c|}{mIoU} & RMSE \\ 
        \cline{2-4}
        \multicolumn{1}{c|}{}    & Vaihingen & Bing huts & Bing huts \\ 
        \hline
        DSAC & 0.825 & 0.617 & 17.5 \\ 
        \hline
        DARNet & 0.871 & 0.758 & 16.9 \\ 
        \hline
        Our model & \textbf{0.911} & \textbf{0.838} & \textbf{12.2} \\ 
        \hline
    \end{tabular}
    \caption{Results on two datasets, reported as mIoU (both Vaihingen and Bing huts) and RMSE (only \textit{Bing huts}).}
    \label{tab:baseline}
\end{table}

\begin{table}[ht]
    \centering
    \begin{tabular}{|c|c|c|c|}
        \cline{3-4}
        \multicolumn{2}{c|}{} & \multicolumn{2}{c|}{mIoU} \\
        \hline
        Method & Backbone & Vaihingen & Bing huts \\
        \hline
        DARNet & DRN & 0.871 & 0.758 \\
        \hline
        DARNet & DRN+ViT-B & 0.904 & 0.815 \\
        \hline
        DARNet & Res50+ViT-B & 0.900 & 0.823 \\
        \hline
        Triple-Branch & DRN+ViT-B & \textbf{0.911} & \textbf{0.838} \\
        \hline
    \end{tabular}
    \caption{Results of our proposed method compared to DARNet using different CNN backbones and model architecture.}
    \label{tab:ablation}
\end{table}

By the ablation experiment we evaluated the effectiveness of each part of our model and the result is shown in Table \ref{tab:ablation}. We can see that by adding a ViT module after dilated residual network the mIoU is significantly improved. And comparing to the DARNet decoder which only used a single upsampling branch, our decoder is more robust and performs better.

\begin{figure*}[htb]
    \centering
        \begin{minipage}{0.13\linewidth}
            \centerline{\includegraphics[width=0.83in]{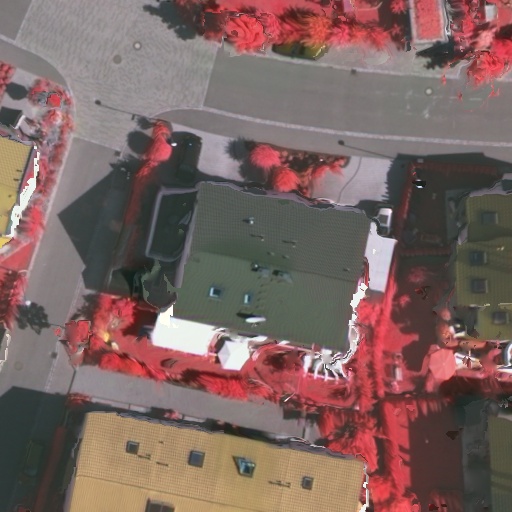}}
            \vspace{3pt}
            \centerline{\includegraphics[width=0.83in]{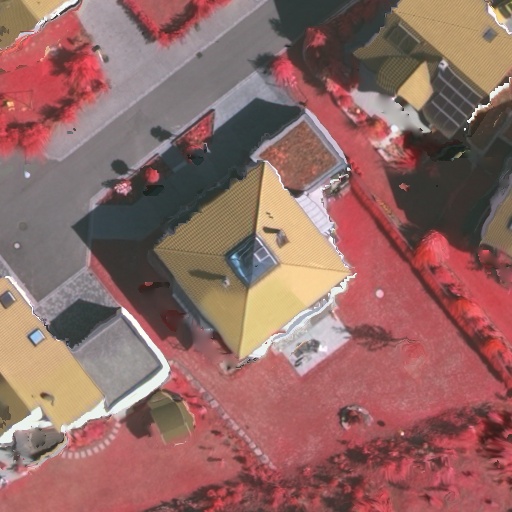}}
            \vspace{3pt}
            \centerline{\includegraphics[width=0.83in]{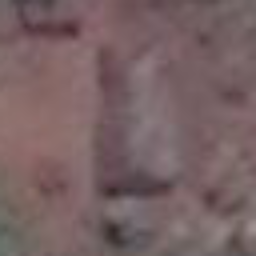}}
            \vspace{3pt}
            \centerline{\includegraphics[width=0.83in]{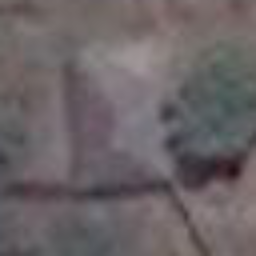}}
            \centerline{(a)}
        \end{minipage}
        \begin{minipage}{0.13\linewidth}
            \centerline{\includegraphics[width=0.83in]{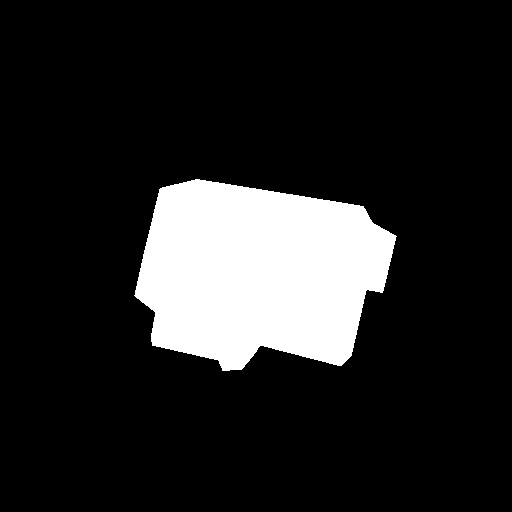}}
            \vspace{3pt}
            \centerline{\includegraphics[width=0.83in]{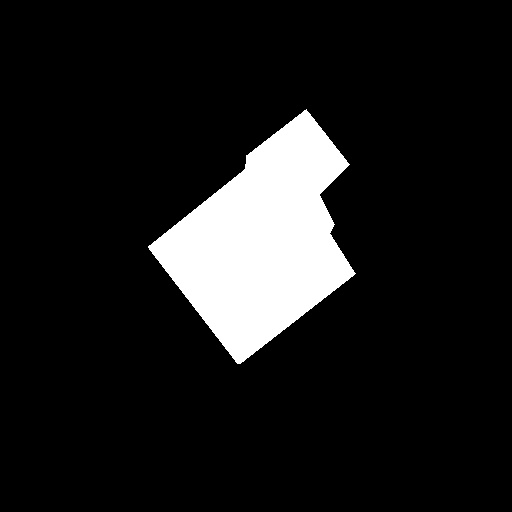}}
            \vspace{3pt}
            \centerline{\includegraphics[width=0.83in]{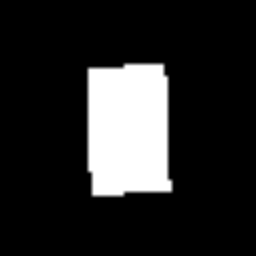}}
            \vspace{3pt}
            \centerline{\includegraphics[width=0.83in]{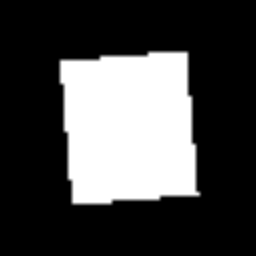}}
            \centerline{(b)}
        \end{minipage}
        \begin{minipage}{0.13\linewidth}
            \centerline{\includegraphics[width=0.83in]{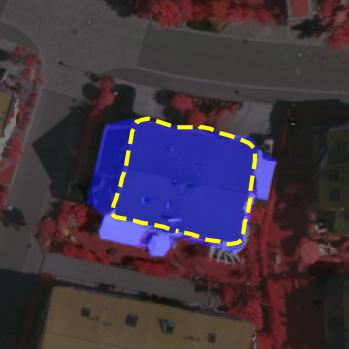}}
            \vspace{3pt}
            \centerline{\includegraphics[width=0.83in]{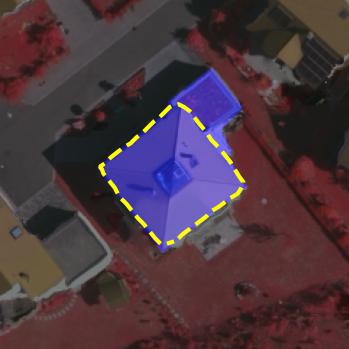}}
            \vspace{3pt}
            \centerline{\includegraphics[width=0.83in]{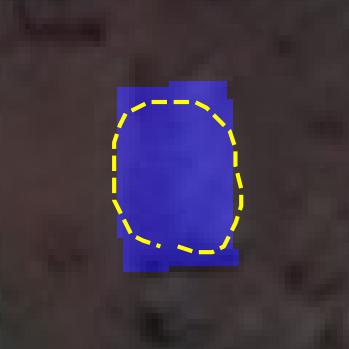}}
            \vspace{3pt}
            \centerline{\includegraphics[width=0.83in]{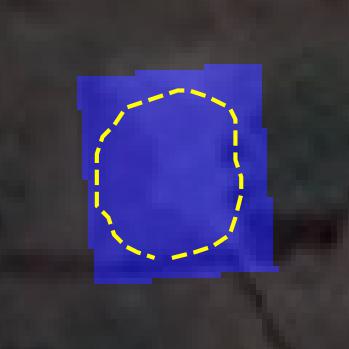}}
            \centerline{(c)}
        \end{minipage}
        \begin{minipage}{0.13\linewidth}
            \centerline{\includegraphics[width=0.83in]{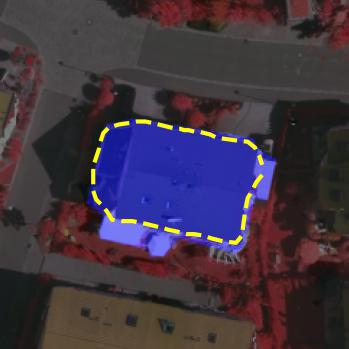}}
            \vspace{3pt}
            \centerline{\includegraphics[width=0.83in]{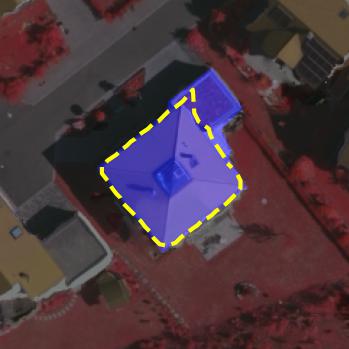}}
            \vspace{3pt}
            \centerline{\includegraphics[width=0.83in]{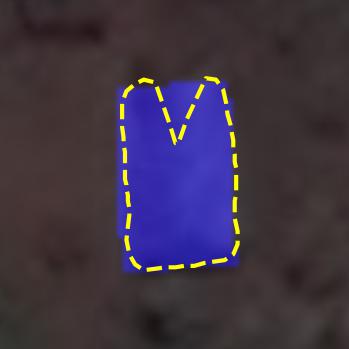}}
            \vspace{3pt}
            \centerline{\includegraphics[width=0.83in]{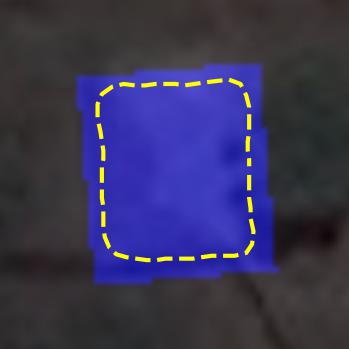}}
            \centerline{(d)}
        \end{minipage}
        \begin{minipage}{0.13\linewidth}
            \centerline{\includegraphics[width=0.83in]{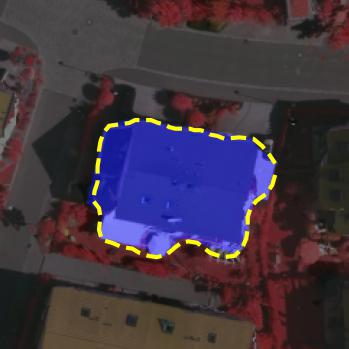}}
            \vspace{3pt}
            \centerline{\includegraphics[width=0.83in]{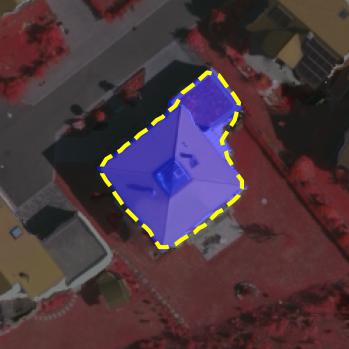}}
            \vspace{3pt}
            \centerline{\includegraphics[width=0.83in]{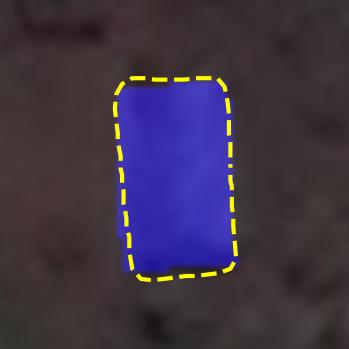}}
            \vspace{3pt}
            \centerline{\includegraphics[width=0.83in]{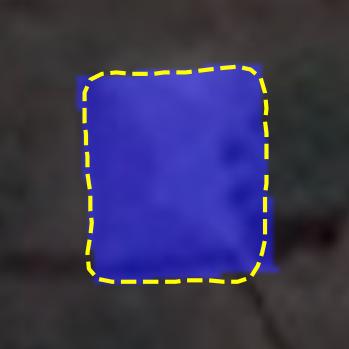}}
            \centerline{(e)}
        \end{minipage}
    \label{fig:visual}
    \caption{Visualization of some prediction samples. (a) is the original image, (b) is ground truth, (c), (d), (e) is the results of DSAC, DARNet and our model. Ground truth in results is shown in blue, and predictions are shown in yellow.}
\end{figure*}

\section{Conclusion}
\label{sec:4}

In this paper we proposed a CNN and Transformer based model to achieve automatic building contour extraction from high-resolution aerial images. In this model we used a hybrid encoder and a triple-branch decoder to deal with different features generated by different part of encoder. The experiment results showed that our model outperformed baselines models on both Vaihingen dataset and Bing huts dataset. And the ablation experiment result validated the effectiveness of the ViT module and the triple-branch structure in our model.

\vfill
\pagebreak







\bibliographystyle{IEEEbib}
\bibliography{strings,refs}

\begin{thebibliography}{10}

\bibitem{ref5}
Michael Kass, Andrew Witkin, and Demetri Terzopoulos,
\newblock ``Snakes: Active contour models,''
\newblock {\em International journal of computer vision}, vol. 1, no. 4, pp.
  321--331, 1988.

\bibitem{ref6}
Xu~Chen, Bryan~M Williams, Srinivasa~R Vallabhaneni, Gabriela Czanner, Rachel
  Williams, and Yalin Zheng,
\newblock ``Learning active contour models for medical image segmentation,''
\newblock in {\em Proceedings of the IEEE/CVF conference on computer vision and
  pattern recognition}, 2019, pp. 11632--11640.

\bibitem{ref7}
Diego Marcos, Devis Tuia, Benjamin Kellenberger, Lisa Zhang, Min Bai, Renjie
  Liao, and Raquel Urtasun,
\newblock ``Learning deep structured active contours end-to-end,''
\newblock in {\em Proceedings of the IEEE Conference on Computer Vision and
  Pattern Recognition}, 2018, pp. 8877--8885.

\bibitem{ref8}
Dominic Cheng, Renjie Liao, Sanja Fidler, and Raquel Urtasun,
\newblock ``Darnet: Deep active ray network for building segmentation,''
\newblock in {\em Proceedings of the IEEE/CVF Conference on Computer Vision and
  Pattern Recognition}, 2019, pp. 7431--7439.

\bibitem{ref9}
Joachim Denzler and Heinrich Niemann,
\newblock ``Active rays: Polar-transformed active contours for real-time
  contour tracking,''
\newblock {\em Real-Time Imaging}, vol. 5, no. 3, pp. 203--213, 1999.

\bibitem{ref10}
Fisher Yu, Vladlen Koltun, and Thomas Funkhouser,
\newblock ``Dilated residual networks,''
\newblock in {\em Proceedings of the IEEE conference on computer vision and
  pattern recognition}, 2017, pp. 472--480.

\bibitem{ref11}
Ashish Vaswani, Noam Shazeer, Niki Parmar, Jakob Uszkoreit, Llion Jones,
  Aidan~N Gomez, {\L}ukasz Kaiser, and Illia Polosukhin,
\newblock ``Attention is all you need,''
\newblock {\em Advances in neural information processing systems}, vol. 30,
  2017.

\bibitem{ref12}
Alexey Dosovitskiy, Lucas Beyer, Alexander Kolesnikov, Dirk Weissenborn,
  Xiaohua Zhai, Thomas Unterthiner, Mostafa Dehghani, Matthias Minderer, Georg
  Heigold, Sylvain Gelly, et~al.,
\newblock ``An image is worth 16x16 words: Transformers for image recognition
  at scale,''
\newblock {\em arXiv preprint arXiv:2010.11929}, 2020.

\bibitem{ref13}
Jieneng Chen, Yongyi Lu, Qihang Yu, Xiangde Luo, Ehsan Adeli, Yan Wang, Le~Lu,
  Alan~L Yuille, and Yuyin Zhou,
\newblock ``Transunet: Transformers make strong encoders for medical image
  segmentation,''
\newblock {\em arXiv preprint arXiv:2102.04306}, 2021.

\bibitem{ref14}
Hu~Cao, Yueyue Wang, Joy Chen, Dongsheng Jiang, Xiaopeng Zhang, Qi~Tian, and
  Manning Wang,
\newblock ``Swin-unet: Unet-like pure transformer for medical image
  segmentation,''
\newblock {\em arXiv preprint arXiv:2105.05537}, 2021.

\end{thebibliography}

\end{document}